\documentclass[letterpaper, 10 pt, conference]{ieeeconf}  

\IEEEoverridecommandlockouts                              

\usepackage{hyperref}
\usepackage{graphics} 
\usepackage{epsfig} 
\usepackage{mathptmx} 
\usepackage{times} 
\usepackage{amsmath} 
\usepackage{amssymb}  
\usepackage{xcolor}
\usepackage{multirow}
\usepackage{booktabs}
\usepackage{siunitx}
\usepackage[format=plain]{caption}
\usepackage{cite}

\usepackage{subcaption}
\let\labelindent\relax
\usepackage{enumitem}

\usepackage{hyperref}
\usepackage{xcolor}
\usepackage{comment}

\title{\LARGE \bf
An End-to-End Framework of Road User Detection, Tracking, and Prediction from Monocular Images
}

\author{Hao Cheng$^{1}$, Mengmeng Liu$^{2}$, and Lin Chen$^{3}$ 
\thanks{$^{1}$Scene Understanding Group, University of Twente, The Netherlands,
        {\tt\small h.cheng-2@utwente.nl,} Cheng is funded by MSCA European
Postdoctoral Fellowships under the 101062870 - VeVuSafety project.}%
\thanks{$^{2}$Institute of Cartography and Geoinformatics, Leibniz University Hannover, Germany,
    {\tt\small mengmeng.liu1998@gmail.com.}}%
\thanks{$^{3}$VISCODA GmbH,
        Schneiderberg 32, 30167 Hannover, Germany,
        {\tt\small chen@viscoda.com.}}%
}

\begin{document}

\maketitle
\thispagestyle{empty}
\pagestyle{empty}

\begin{abstract}
Perception that involves multi-object detection and tracking, and trajectory prediction are two major tasks of autonomous driving. 
However, they are currently mostly studied separately, which results in most trajectory prediction modules being developed based on ground truth trajectories without taking into account that trajectories extracted from the detection and tracking modules in real-world scenarios are noisy. 
These noisy trajectories can have a significant impact on the performance of the trajectory predictor and can lead to serious prediction errors. 
In this paper, we build an end-to-end framework for detection, tracking, and trajectory prediction called ODTP (Online Detection, Tracking and Prediction). 
It adopts the state-of-the-art online multi-object tracking model, QD-3DT, for perception and trains the trajectory predictor, DCENet++, directly based on the detection results without purely relying on ground truth trajectories. 
We evaluate the performance of ODTP on the widely used nuScenes dataset for autonomous driving. 
Extensive experiments show that ODPT achieves high performance end-to-end trajectory prediction. 
DCENet++, with the enhanced dynamic maps, predicts more accurate trajectories than its base model. 
It is also more robust when compared with other generative and deterministic trajectory prediction models trained on noisy detection results. 

\end{abstract}

\section{Introduction}
\label{introduction}
Trajectory prediction plays a crucial role in achieving autonomous driving. 
It involves observing the behavior of agents like vehicles, pedestrians, and other road users in a few past time steps. 
This observation information includes perceiving the road users' type and their past trajectories, which is then fed into a trajectory predictor to forecast their potential trajectories in the following time steps.
Despite the rapid development of trajectory prediction methods, they are usually developed independently of the perception module, assuming that the ground truth information of the road users' past trajectories is already available. 
This means that trajectory predictors trained on ground truth data may not reflect the complexities of real-world driving scenarios \cite{zhang2023towards}.
Moreover, the input data to the trajectory prediction module is prone to noise because the perception module is imperfect due to long-standing issues such as changes in lighting, scale, background interference, sensor limitations, and multiple occlusions.
Therefore, this paper aims to address the task of trajectory prediction by integrating the perception module and developing an end-to-end framework for road user detection, tracking, and prediction.

Object detection results lay as the foundation for multi-object tracking and trajectory prediction. 
Our focus is primarily on mixed traffic scenarios that comprise not only vehicles and pedestrians but also other types of road users.
In this paper, we have selected monocular images obtained from a moving vehicle as the input for the perception module, as camera sensors are low-cost for capturing all the objects in the field of view and straightforward to deploy. 
The perception module we have employed is the monocular Quasi-Dense 3D Object Tracking (QD-3DT) \cite{hu2022monocular} as the base model, which can effectively associate moving agents over time and estimate their complete 3D bounding box information from a sequence of 2D images captured on a mobile platform.

To achieve end-to-end prediction, the prediction module takes the 2D positional information at discrete time steps from the perception module as input. 
In order to consider the multimodality nature of agents' movements and their mutual influence during interactions, we use the DCENet model \cite{cheng2021exploring} as the base model for the prediction module.
DCENet explores the spatial and temporal information captured by dynamic maps and leverages the attention mechanisms \cite{vaswani2017attention} and Conditional Variational Autoencoder \cite{kingma2014semi} to predict agents' multimodal trajectories. 
Instead of relying on ground truth point-wise trajectory data, we extend DCENet to take as input the perceived trajectory data, including the agents' shape and pose information. 
This approach allows the dynamic maps to more accurately map the position, velocity, and pose information of each agent into the 2D grid cells that are projected from the agents' detected 3D shape.
Furthermore, the extended DCENet model is trained based on detection results, which are more robust against detection noise compared to ground truth data. 
We term this new prediction model DCENet++ to reflect its improved performance.

Figure~\ref{fig:End-to-End} depicts the end-to-end detection, tracking, and prediction framework overview. 
It comprises three primary components: monocular image sensor input captured from the ego vehicle, the QD-3DT-based 3D detection and tracking module, and the DCENet++ multimodal prediction module for trajectory prediction. 
We name our end-to-end framework ODTP (Online Detection, Tracking and Prediction).
The \textbf{key contributions} of our work are as follows:

\begin{figure*}[hbpt!]
\centering
\includegraphics[clip=true, trim=2in 4.5in 3.5in 1.8in,width=1\textwidth]{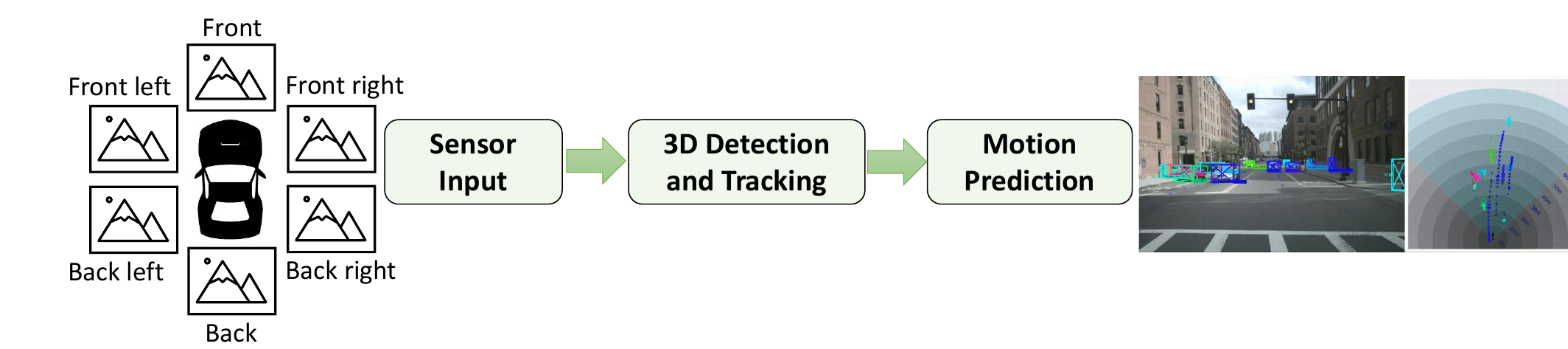}
\vspace{-12pt}
\caption{The overview end-to-end detection, tracking and trajectory prediction framework using camera sensors.}
\vspace{-12pt}
\label{fig:End-to-End}
\end{figure*}

\begin{itemize}
    \item We propose an end-to-end framework ODTP, which automatically detects and tracks various types of road users from monocular images and predicts their multimodal trajectories. 
    The prediction module in ODTP is trained directly on the detection and tracking results, rather than on the manually labeled ground truth data.
    \item We extend the trajectory prediction model DCENet by adding road users' shape and pose information acquired by the detection module, called DCENet++, which achieves better performance than DCENet that only relies on the 2D positional information.     
\end{itemize}

\section{Related Work}
\label{relatedwork}

\subsection{Object Detection and Tracking}
\label{3D Object Tracking}
In this paper, we focus on object detection and tracking based on images.
In the paradigm of object detection, the most attention has been given to two-stage detectors represented by R-CNN-based models with region proposal networks \cite{girshick2014rich,girshick2015fast,ren2015faster} and one-stage detectors represented by YOLO-based models \cite{redmon2016you,redmon2017yolo9000}.
SORT \cite{bewley2016simple} uses Faster R-CNN \cite{ren2015faster} for object detection and predicts and updates the motion track using the Kalman filter \cite{stone2013bayesian}. 
The Hungarian dichotomous matching algorithm \cite{kuhn1955hungarian} with Intersection Over Union (IOU) as the matching criterion is then used to match detected and tracked objects.
DeepSORT \cite{wojke2017simple} further adds appearance representation of detections using deep neural networks to enhance the tracking performance. 
Several works, such as \cite{scheidegger2018mono,sharma2018beyond,luiten2020track,li2018stereo,osep2017combined}, propose the use of 3D information to narrow the search area and make the object's trajectory smoother.
A dominant approach to associating data in multi-object tracking (MOT) problems is to utilize different kinds of costs, including trajectory priors, bounding box center locations, optical flow, bounding box overlap, and appearance information or deep appearance features \cite{sharma2018beyond,lu2020retinatrack,hu2022monocular}. 
For example, \cite{sharma2018beyond} employs simple complementary costs for data association, which include 2D-3D cost, 3D-3D cost, appearance cost, and shape and pose cost.
\cite{lu2020retinatrack} employs discriminative feature embeddings and the greedy bipartite matching method to match new detections and trajectories.
Moreover, DEFT \cite{chaabane2021deft} proposes a joint detection and tracking method that relies on the appearance features from a detection backbone for object-to-track association.
3D-Times \cite{li2022time3d} employs attention mechanisms to learn spatial-temporal information cues for joint 3D detection and tracking from monocular videos.
In this paper, we use QD-3DT \cite{hu2022monocular} as our base model for joint 3D detection and tracking. 
It employs Faster R-CNN for 2D detection basis and Region of Interest (ROI) features for each proposal extracted by a region proposal network to regress the 3D dimensions, 2D projection center, depth, and orientation.
Moreover, different similarity cues, such as the deep representation similarity, the overlap of 3D bounding boxes, and the motion similarity, are utilized for data association. 
In addition, QD-3DT uses an LSTM-based module instead of using 3D KF for motion refinement.

\subsection{Trajectory Prediction}
\label{GAN, CVAE-based trajectory predictor}
Recent years, deep learning methods, such as Generative Adversarial Network (GAN) \cite{goodfellow2014generative}, Conditional Variational AutoEncoder (CVAE) \cite{kingma2014semi,sohn2015learning}, and attention mechanisms \cite{cheng2022gatraj,liu2023laformer}, have been introduced to the trajectory prediction task.
Gupta et al. \cite{gupta2018social} propose a fusion of LSTM and GAN, using the global pooling module of LSTM as the encoder-decoder generator and a discriminator composed of multiple LSTMs. 
For machine navigation, Altan et al. \cite{altan2017glidepath} propose a pedestrian-dependent spatio-temporal graphical representation that can effectively represent the importance of pedestrians in congested environments. 
Social-BiGAT \cite{kosaraju2019social} proposes the Graph Attention Network (GAT) \cite{velivckovic2017graph} to learn feature representations and performs reversible transformations between the scene and the response's underlying noise vector.
Similarly, SR-LSTM \cite{zhang2019sr} employs Graph Neural Networks (GNN) to model the interconnections among agents and predicts their deterministic future trajectories.
AgentFormer \cite{yuan2021agentformer} utilizes the Transformer network \cite{vaswani2017attention} to learn spatial-temporal information of agents and applies CVAE for multimodal trajectory predictions.
In contrast, \cite{cheng2021exploring} proposes a model called DCENet that utilizes self-attention and LSTM to model interactions between agents and a CVAE framework to predict a set of possible trajectories conditioned on its observed trajectory and the learned dynamic context for each agent.
Considering its superior performance, we utilize DCENet as our prediction model baseline. In comparison to the original DCENet, we add the agent's shape and pose information to the dynamic maps, which can model interactions more accurately. 
We also adopt DCENet to moving camera scenes, i.e., real autonomous driving scenarios.
In this way, in contrast to training the prediction model using ground truth trajectories, our model utilizes camera data only, and the prediction is conducted on detection and tracking results.

\subsection{Joint Tracking and Forecasting}
\label{Joint tracking and forecasting}
Recently, a few studies have investigated the possibility of joint MOT and trajectory prediction~\cite{liang2020pnpnet,weng2021ptp,zhang2023towards}.
Weng et al. \cite{weng2021ptp} propose a novel data association method that utilizes GNNs to model interactions between new detections and trajectories. 
After message passing by GNNs, the affinity matrix between new detections and trajectories is learned for association. 
Instead of processing the MOT task first and then the trajectory prediction task, Weng et al. \cite{weng2021ptp} process the two tasks in parallel, so the trajectory prediction task does not explicitly depend on the MOT results.
Liang et al. \cite{liang2020pnpnet} change the order of tasks for MOT and trajectory prediction. 
Unlike the traditional approach, it first carries out joint detection and prediction tasks before updating the object trajectory.
Zhang et al. \cite{zhang2023towards} propose to extract the motion information based on the affinity cues among detection results and predict trajectories directly based on detection results.
Inspired by the query-based end-to-end object detection with transformers \cite{carion2020end}, end-to-end perception and motion prediction is achieved by extending the object query with recurrent temporal information, such as ViP3D \cite{gu2023vip3d} and UniAD \cite{hu2023planning}. 
In contrast to those works, our work utilizes camera data only to build an end-to-end detection, tracking, and prediction framework. 
The trajectories extracted using monocular images have noise and introduce a greater challenge to the task of trajectory prediction. 
As most trajectory prediction methods are trained and inferenced on ground truth trajectories, the effect of noise in the detection and tracking tasks on trajectory prediction is not considered.
To the opposite, we compare the performance differences between generative and deterministic trajectory prediction models when dealing with such noisy trajectories.

\section{Methodology}
\label{methodology}
\subsection{Problem Formulation}
\label{Problem Formulation}
The goal of the end-to-end framework ODPT is to take a monocular image sequence as input for the 3D object detection module and output a set of 3D bounding boxes (bbx) at frame $t$, denoted as $S=\{s^t_{1},...,s^t_{J}\}$.
After performing data association and motion refinement in the MOT module, which takes the bbx as input, we obtain a series of smooth trajectories denoted as $\mathbb{T}=\{\tau_{1},...,\tau_{N}\}$, where $\tau_{i} \in \mathbb{R}^{T\times 2}$, and the refined 3D bounding boxes denoted as $\mathbb{S}=\{s^t_{1},...,s^t_{N}\}$. 
Here, $i \in \{1, ..., N\}$, $N \leq J$ represents the total number of detected and tracked agents in the given scenario, and $T$ is the observed time horizon. $T\geq 2$, to make sure that we have enough observed steps to derive the speed and pose information.
Subsequently, the trajectories $\mathbb{T}$ and detection bbx $\mathbb{S}$ serve as the input to the trajectory prediction module. 
In this module, we predict a set of possible future trajectories denoted as $\{\hat{Y}^{T+1:T'}_{i,1},...,\hat{Y}^{T+1:T'}_{i,K}\}$ conditioned on the detected trajectories and bbx for each agent $i$. 
Here, $K$ represents the number of predicted trajectories and $T'-T$ represents the predicted time horizon.
In the following sections, we explain each module of ODPT in detail.

\subsection{QD-3DT}
\label{Monocular Quasi-Dense 3D Object Tracking}
The goal of the QD-3DT module \cite{hu2022monocular} is to provide the 3D information of all tracked objects by inputting consecutive frames of monocular images and GPS/IMU information from the ego vehicle. 
The GPS/IMU data is used to obtain localization information about the ego vehicle's motion.
To achieve this, we transfer the 3D information, including the shape, pose, and position of all neighboring agents, from the cameras to the local frame of the ego agent. 
During the perception process, the monocular images are first processed through a backbone network, such as VGG16 \cite{simonyan2014very}, and a Region Proposal Network (RPN) \cite{ren2015faster}, to generate 2D regions of interest (ROIs). 
These ROIs are then fed into two multi-head networks, which output similarity feature embeddings and 3D layouts.
To track 3D object instances over time, multimodal similarity metrics between the tracked trajectories and detected objects are computed utilizing 3D information, motion information, and feature embeddings. Additionally, motion-aware data association and depth-ordering matching techniques are used to mitigate occlusion problems. 
Finally, the tracking module refines the 3D information of the objects.
It is worth noting that in our approach, we directly use the image pixel coordinates without normalization to calculate the IOU between the detected and ground truth bbx. 
This differs from the original setting of QD-3DT. We made this choice based on empirical findings that using normalized coordinates changes the image scale and leads to larger detection errors \cite{hu2022monocular}.

\subsection{DCENet++}
\label{Dynamic Maps V2}

\begin{figure}[t!]
\centering
\includegraphics[clip=true, trim=0cm 1cm 0cm 1cm, width=1\linewidth]{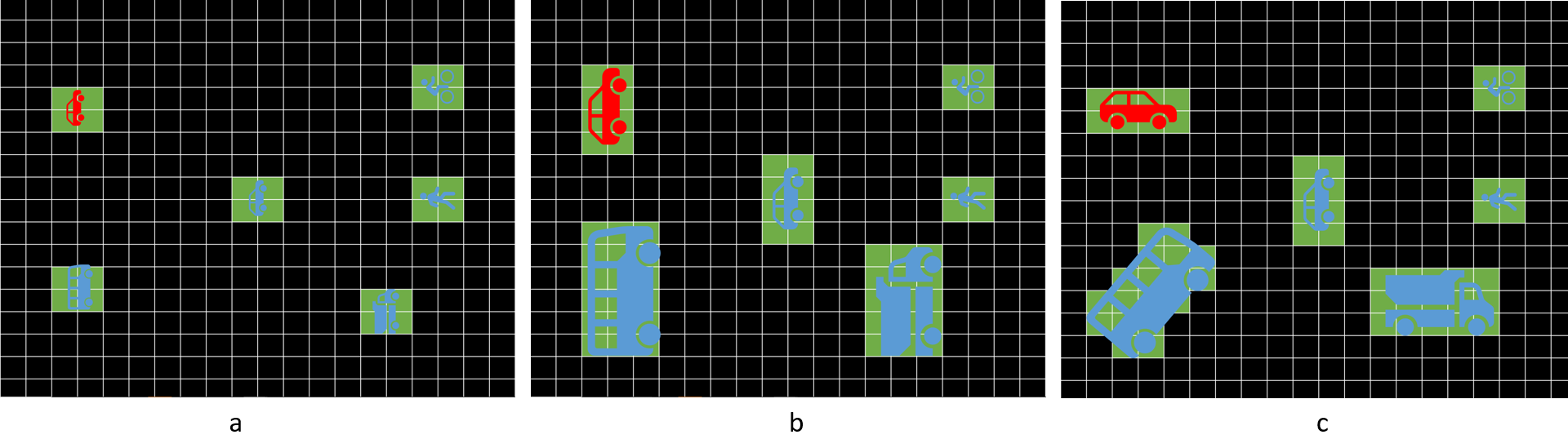}
\caption{Comparison between the original and the refined dynamic maps with the agents' shape and pose information. (a) No agents' shape and pose information (b). Only considering the agents' shape information, and (c) Considering both the agents' shape and pose information.}
\label{fig:DM V2}
\vspace{-12pt}
\end{figure}

We utilize the DCENet model \cite{cheng2021exploring} for the trajectory prediction module and adapt it from a bird's-eye view to the ego perspective of the mobile cameras. 
With the 3D tracking module from QD-3DT, we not only estimate the 3D object center $\{x,y,z\}$ but also the object dimensions $D=\{l,h,w\}$ and object pose $\theta$ for each agent.
Therefore, compared to the original dynamic maps that use an approximation of the agent's shape, we use the detected shape and pose information to refine the dynamic maps, allowing for a more accurate modeling of the neighboring agents' position, velocity, and pose information. 
In this work, we assume that all agents are moving on the ground surface and our prediction task is focused on the 2D positions of the $x$- and $y$-coordinates. 
Consequently, the dynamic maps for all the objects are modelled based on the projection on the ground plane, leaving object height $h$ and altitude $z$ out of consideration.
Fig.~\ref{fig:DM V2} (a) shows the original dynamic map \cite{cheng2021exploring}, which assumes that each agent has the same size and orientation. 
This approximation is based on the observation that pedestrian size varies little within the pedestrian trajectory dataset (e.g., ~\cite{pellegrini2009you}). 
Since pedestrians are relatively small and occupy only one grid cell in the dynamic maps, orientation information can be disregarded in the pedestrian dataset.
However, when adapting DCENet to autonomous driving scenarios, we need to consider the significant shape differences among heterogeneous types of agents, such as vehicles and cyclists. 
To ensure proper alignment, especially for large agents occupying multiple grid cells in the dynamic maps, the occupied grid cells should align with the agent's pose.
Fig.~\ref{fig:DM V2} (b) and (c) illustrate the dynamic maps that include shape information alone and both shape and pose information, respectively. 
These refined dynamic maps enable us to handle objects with varying shapes and poses. We refer to DCENet with the refined dynamic maps as DCENet++.

\subsection{Joint 3D Tracking and Forecasting}
\label{Joint 3D tracking and forecasting}
With the QD-3DT perception module, which includes data association and tracking, we obtain a set of tracked trajectories $\mathbb{T}=\{\tau^{1:T}_{1},...,\tau^{1:T}_{N}\}$ from previous frames, as well as a set of detections $\mathbb{S}=\{s^T_{1},...,s^T_{N}\}$ at frame $T$.
We configure the batch size of DCENet++ to match the number of objects detected in the current frame, the same as $N$. 
This choice allows us to focus solely on the interactions among agents that are present concurrently and successfully detected in the given frames.
To model the interactions between the ego agent and its neighboring agents, we employ the extended dynamic maps. 
At each time step $t\in \{1, ..., T\}$, we project the neighboring agents onto the grid cells of the dynamic map, centered at the current position $\{x, y\}$ of the ego agent. The projection is based on the detected 3D shape $\{l,h,w\}$ obtained from $\mathbb{S}$. 
Next, we map the position, velocity, and pose information derived from the tracked trajectories $\mathbb{T}$ onto dedicated channels in the dynamic maps, following the approach described in \cite{cheng2021exploring}.
Simultaneously, the offset sequence $\Delta X^{1:T-1}_{i}=\{\Delta x^{1}_{i},...,\Delta x^{T-1}_{i}\}$ $\in$ $\mathbb{R}^{(T-1) \times 2 }$ for each agent's trajectory is combined with the sequence of dynamic maps at the corresponding time steps. 
These combined inputs serve as the joint condition for the prediction module of DCENet++.
Finally, DCENet++ predicts multimodal trajectories $\{\hat{Y}^{T+1:T'}_{i,1},...,\hat{Y}^{T+1:T'}_{i,K}\}$ for all agents.

\section{Experiment}
\label{experiment}

\subsection{Dataset}
\label{dataset}
We evaluate our ODTP on nuScenes~\cite{caesar2020nuscenes}, which is one of the most commonly used large-scale real-word datasets for autonomous driving.
The ego vehicle is equipped with multiple sensors, such as LiDAR, monocular camera, and radar. 
In this paper, we focus on the camera images.
In total, the dataset contains 1000 driving scenes in Boston and Singapore, including 700 scenes for training, 150 scenes for validation and 150 scenes for test. 

\subsection{Evaluation Metrics}
\label{metrics}

\noindent\textbf{MOT metrics.}
\label{Multiple Object Tracking Metrics}
We adhere to the tracking metrics established by nuScenes, specifically AMOTA and AMOTP \cite{weng2019baseline}.
AMOTA stands for averaged multi-object tracking accuracy (MOTA) \cite{bernardin2006multiple} at various recall thresholds. 
MOTA provides a comprehensive assessment by taking into account false positives, missed targets, and identity switches.
AMOTP refers to the average multi-object tracking precision (MOTP) \cite{bernardin2006multiple}. 
MOTP quantifies the misalignment between the annotated and predicted bounding boxes, offering insights into the accuracy of object localization.

\noindent\textbf{Trajectory prediction metrics.}
\label{Trajectory Prediction Metrics}
We employ two widely used metrics to evaluate the trajectory prediction task: Average Displacement Error (ADE) and Final Displacement Error (FDE) \cite{alahi2016social}. 
ADE calculates the average Euclidean distance between the predicted trajectory and the corresponding ground truth trajectory, while FDE measures the Euclidean distance between their final positions.
Consistent with previous works \cite{gupta2018social,chai2019multipath,sadeghian2019sophie}, we select the minimum ADE ($\text{ADE}_{K}$) and FDE ($\text{FDE}_{K}$) from the best prediction among $K$ trajectory samples for each agent.

\subsection{Experimental Setting} 
\label{Setting}
Following the VeloLSTM module in QD-3DT \cite{hu2022monocular}, we set the observation time horizon to 2.5 seconds with a frame rate of 2 Hz. 
The prediction time horizon is set to 4 seconds with the same frame rate.

For the 3D detection and tracking module, we adopt the approach presented in QD-3DT \cite{hu2022monocular}. 
We utilize a pretrained Faster R-CNN \cite{ren2015faster} model on ImageNet \cite{russakovsky2015imagenet} from TorchVision \cite{paszke2019pytorch} for 2D detection and 3D center estimation. 
However, to address accumulated tracking errors, we modify the default setting in QD-3DT. 
Instead of continuously predicting the object state until it goes beyond the tracking range or its lifespan ends (e.g., ten time steps), we use the predicted bbx at the next step once to compute the affinity between the trajectory and the detected object state.
Moreover, different from the default setting in QD-3DT, we use the image pixel coordinates without normalization to calculate the IOU between the detected and ground truth bbx.

During training, the VeloLSTM module is trained for 100 epochs using ten sample frames per object trajectory, with a batch size of 128. 
For trajectory prediction model DCENet++, we employ the Adam optimizer \cite{kingma2014adam} with early stopping, setting the patience parameter to ten to prevent overfitting. 
The initial learning rate is set to $10^{-4}$, and we decay the learning rate by a factor of 0.5 every 20 epochs.

\section{Results} 
\label{results}
In this section, we begin by evaluating the performance of the perception module. 
Subsequently, we assess the performance of the trajectory prediction module. 
Finally, we present the qualitative performance of the ODPT framework for end-to-end object detection, tracking, and prediction.

\subsection{Perception Performance}
Firstly, we compare our setting with the default setting in terms of computing the IOU between the detected bounding box and the ground truth annotation, as well as the affinity between the trajectory and the detected object state.
Table~\ref{table:Comparisonoptimization} illustrates the results of this comparison. 
The improved AMOTA and the decreased AMOTP demonstrate that our pixel-based IOU computation and the one-time update of the detected object state yield better performance compared to the default setting in QD-3DT.

\begin{table}[ht]
\centering
\caption{Perception performance for multi-object tracking.}
\begin{tabular}{cc|cc} 
\toprule
IOU  & state update & \textbf{AMOTA}$\uparrow$ & \textbf{AMOTP}$\downarrow$ \\ \midrule
default & default & 0.233 & 1.528 \\
pixel & default & 0.235  & 1.517 \\
pixel & one time & \textbf{0.243} & \textbf{1.512} \\ \bottomrule
\end{tabular}
\label{table:Comparisonoptimization}
\end{table}

Next, we explore different IOU thresholds to strike a balance between precision and recall for object detection.
As depicted in Table~\ref{table:ComparisonofIoU.}, we observe that an IOU threshold of 0.5 yields slightly better results compared to the other thresholds. 
Consequently, we adopt an IOU threshold of 0.5 as the default setting for subsequent experiments.

\begin{table}[ht]
\centering
\caption{Comparison of IOU.}
\begin{tabular}{c|cc} 
\toprule
IOU & \textbf{AMOTA}$\uparrow$ & \textbf{AMOTP}$\downarrow$ \\
\midrule
0.4   &  0.236 & 1.516 \\
0.5   & \textbf{0.243}  & \textbf{1.512} \\
0.7 & \textbf{0.243} & 1.515 \\
\bottomrule
\end{tabular}
\label{table:ComparisonofIoU.}
\end{table}

\subsection{Trajectory Prediction Performance}
\label{trajectoryperformance}
We conduct experiments by varying the dimensions of the latent variable in the CVAE-based DCENet++ model.
As presented in Table \ref{table:Ablationlatentvariable}, increasing the dimension from 2 to 32 leads to a reduction in trajectory prediction errors, as measured by $ADE_{10}$ and $FDE_{10}$. 
However, when we further increase the dimension to 64, the prediction performance deteriorates.
Therefore, for subsequent experiments, we maintain a fixed dimension of 32 for the latent variable.

\begin{table}[ht]
\centering
\caption{Different dimensions of latent variable $z$.} 
\vspace{-4pt}
\begin{tabular}{c|cc} 
\toprule
Methods &  $ADE_{10}$$\downarrow$ & $FDE_{10}$$\downarrow$\\
\midrule
$z_{dim}=2$ & 0.82 & 1.54 \\
$z_{dim}=32$ & \textbf{0.79}  & \textbf{1.50} \\
$z_{dim}=64$ & 0.82 & 1.53 \\
\bottomrule
\end{tabular}
\label{table:Ablationlatentvariable}
\end{table}

Additionally, we perform an ablation study on the dimension and pose information in the dynamic maps of DCENet++.
Comparing it to the baseline model, DCENet, which lacks dimension and pose information for aligning agents in the dynamic maps, the performance of DCENet++ is better in terms of both ADE and FDE, as shown in Table \ref{table:Ablation studies on nuScenes for Dynamic Maps.}.
Interestingly, we discover that utilizing either dimension or pose information alone does not result in a clear improvement in performance.
This is because either the lack of dimension or pose information could lead to sub-optimal alignments in the dynamic maps.

\begin{table}[ht]
\centering
\caption{The ablation study for the dynamic maps.}
\begin{tabular}{c|cc|cc} 
\toprule
Methods& dimension & pose & $ADE_{10}$$\downarrow$ & $FDE_{10}$$\downarrow$\\
\midrule
DCENet   & - & - & 0.80 & 1.51 \\
DCENet+  & $\surd$ & - & 0.80  & 1.51 \\
DCENet+  & - & $\surd$  & 0.80  & \textbf{1.50} \\
DCENet++  & $\surd$ & $\surd$ & \textbf{0.79} & \textbf{1.50} \\
\bottomrule
\end{tabular} 
\vspace{-8pt}
\label{table:Ablation studies on nuScenes for Dynamic Maps.}
\end{table}

After adjusting the hyperparameters of DCENet++, we conducted a performance comparison with two well-known trajectory prediction models using the nuScenes motion prediction dataset, as shown in Table \ref{table:Evaluation trajectory prediction generative.}.
It should be noted that, for a fair comparison, all these models do not use any map with scene context information.
One of the models we compared DCENet++ to is AgentFormer \cite{yuan2021agentformer}, which is a transformer and CVAE-based model capable of generating multimodal predictions for each agent. 
In the multimodal prediction task, DCENet++ outperforms AgentFormer in predicting both five and ten modalities.
Furthermore, DCENet++ exhibits superior performance compared to AgentFormer in single-modal prediction as well.
Given these promising results, we proceeded to compare DCENet++ with the deterministic model SR-LSTM \cite{zhang2019sr}, which utilizes GNN to model interactions among agents. 
In terms of both ADE and FDE, DCENet++ surpasses SR-LSTM.

\begin{table}[hbpt!]
\centering
\caption{Evaluation of trajectory prediction on nuScenes.}
\setlength{\tabcolsep}{2.8pt}
\begin{tabular}{c|cc|cc|cc} 
\toprule
Methods& $ADE_{1}$$\downarrow$& $FDE_{1}$$\downarrow$& $ADE_{5}$$\downarrow$& $FDE_{5}$$\downarrow$& $ADE_{10}$$\downarrow$& $FDE_{10}$$\downarrow$\\
\midrule
AgentFormer \cite{yuan2021agentformer} & 7.91 & 4.55 & 1.67& 2.62& 1.06 & 1.56 \\
SR-LSTM~\cite{zhang2019sr} & 1.29 & 2.57 & - &  - & -  & - \\
DCENet++ & \textbf{0.97} & \textbf{1.86}& \textbf{0.86} & \textbf{1.62} &\textbf{0.79} & \textbf{1.50} \\
\bottomrule
\end{tabular}
\label{table:Evaluation trajectory prediction generative.}
\end{table}

In the following analysis, we examine the impact of noisy MOT training data by comparing the performance of DCENet++ when trained on ground truth (GT) trajectories versus MOT trajectories.
The results presented in Table \ref{tab:GTandMOT} clearly demonstrate that, as anticipated, training the prediction model DCENet++ using GT trajectories and subsequently testing it on MOT trajectories leads to a significant drop in performance.
Namely, the prediction errors increases by over 200\% when compared to the realistic setting of both training and testing DCENet++ using the MOT trajectory data.
This indicates that a model trained solely on GT trajectories may struggle to generalize effectively in real-world driving scenarios, where the perception module unavoidably produces noisy trajectories during the observation period.
However, this issue can be largely mitigated when DCENet++ is trained and tested on MOT trajectories. 
Remarkably, the MOT-trained DCENet++ exhibits impressive generalization capabilities when deployed in testing on GT trajectories, performing only slightly worse than the ideal scenario where both training and testing are conducted on GT trajectories.

\begin{table}[hbpt!]
\centering
\caption{Comparison between using ground truth and MOT trajectory data. Using the MOT data for both training and testing is referred as the baseline of prediction errors.}
\vspace{-6pt}
\setlength{\tabcolsep}{4.5pt}
\begin{tabular}{l|ll|ll|ll}
\toprule
\multicolumn{1}{c|}{\multirow{2}{*}{Model}} & \multicolumn{2}{c|}{Training}                     & \multicolumn{2}{c|}{Testing}                    & \multicolumn{2}{c}{Errors}                        \\ 
\multicolumn{1}{c|}{}                       & \multicolumn{1}{c}{MOT} & \multicolumn{1}{c|}{GT} & \multicolumn{1}{c}{MOT} & \multicolumn{1}{c|}{GT} & \multicolumn{1}{c}{$ADE_{10}$$\downarrow$} & \multicolumn{1}{c}{$FDE_{10}$$\downarrow$} \\ \midrule
DCENet++                                   &                         &  $\surd$               &  $\surd$                &                        & 2.54 {\color{magenta}(+222\%)}                    & 4.59 {\color{magenta}(206\%)}                   \\ 
DCENet++                                   &  $\surd$                &                        &  $\surd$                &                        & 0.79                    & 1.50                    \\ 
DCENet++                                   &  $\surd$                &                        &                         &   $\surd$              & 0.47 {\color{green}(-40\%)}                   & 0.94  {\color{green}(-37\%)}                  \\ 
DCENet++                                   &                         &  $\surd$               &                         &  $\surd$               & 0.44 {\color{green}(-44\%)}                   & 0.90  {\color{green}(-40\%)}                 \\ \bottomrule
\end{tabular}
\label{tab:GTandMOT}
\vspace{-8pt}
\end{table}

\begin{figure*}[t!]
\centering
\includegraphics[clip=true, trim=0.3cm 0.2cm 0.15cm 0.3cm, , width=1\textwidth]{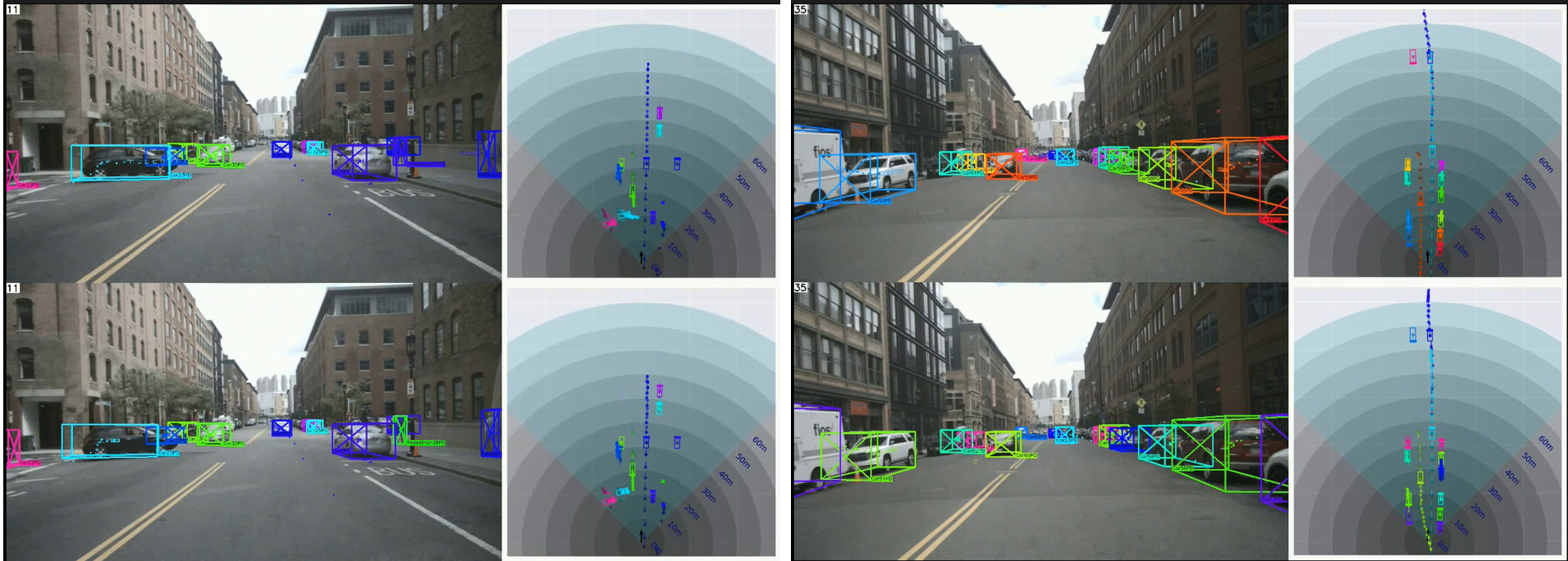}
\caption{Qualitative results of DCENet++ (first row) and SR-LSTM~\cite{zhang2019sr} (second row). On the left is the ego-driving perspective and on the right is the bird's-eye view. The squares represent predicted trajectories and circles denote history trajectories.}
\label{fig:DCE_SR}
\end{figure*}

\begin{figure*}[t!]
\centering
\includegraphics[width=1\textwidth]{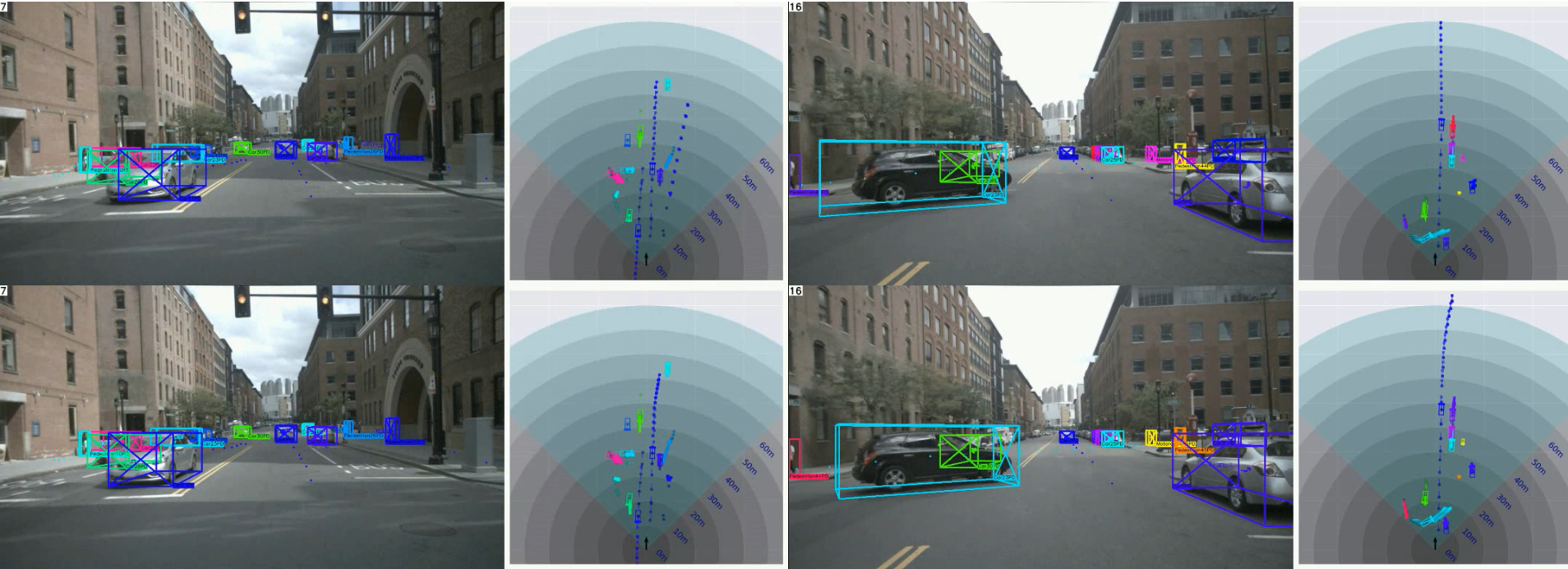}
\caption{Qualitative results of DCENet++ (first row) and AgentFormer~\cite{yuan2021agentformer} (second row). On the left is the ego-driving perspective and on the right is the bird's-eye view. The squares represent predicted trajectories and circles denote history trajectories.}
\label{fig:DCE_VS_AGENTFORMER}
\end{figure*}

\begin{figure*}[t!]
\centering
\includegraphics[width=1\textwidth]{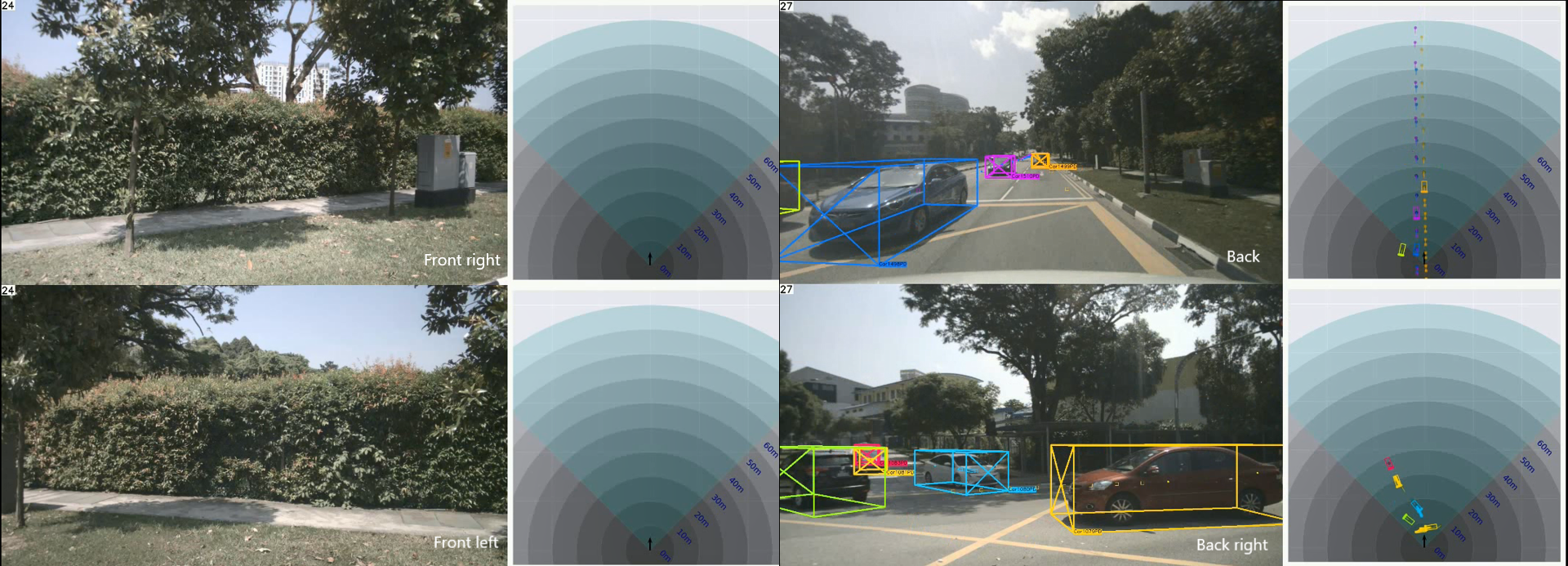}
\caption{Qualitative results of DCENet++ from multi-camera perspectives. On the left is the ego-driving perspective and on the right is the bird's-eye view. The squares represent predicted trajectories and circles denote history trajectories.}
\label{fig:Mutiview1}
\vspace{-12pt}
\end{figure*}

\subsection{Qualitative Results}
\label{Qualitativeresults}

We present the performance analysis of the ODPT in various driving scenarios. 
To visualize the results, we employ the visualization method proposed by Hu et al. \cite{hu2022monocular}.
Specifically, we display the detection results obtained from the camera view and plot the tracked and predicted trajectories using a bird's-eye view perspective above the ego vehicle. 
Fig. \ref{fig:DCE_SR} and \ref{fig:DCE_VS_AGENTFORMER} showcase the outputs of DCENet++ and SR-LSTM, and DCENet++ and AgentFormer models, respectively, revealing that they generate realistic predictions. 
However, compared to the other models, the predictions from DCENet++ exhibit smoother trajectories even though the input trajectory data from the MOT module is noisy.

Furthermore, Fig. \ref{fig:Mutiview1} demonstrates the adaptability of ODPT to monocular images captured from various perspectives. 
This demonstrates the versatility and effectiveness of the ODPT approach across different camera viewpoints.

\vspace{6pt}
\noindent\textbf{Limitations.}
In spite of the promising performance demonstrated above, in this work, we need to separately train the perception module and the trajectory prediction module, which is time-consuming.
Also, because they are separately trained, we could not share the intermediate feature maps to unify the encodings for both the perception and trajectory prediction tasks.
Moreover, whether the multimodal trajectory prediction in this work can consolidate the data association in MOT through providing a more time and space consistent object motion could be further explored.
We leave this as our future work.
Last but not least, when the perception model mis-detects agents due to, e.g., occlusions and lighting conditions, the trajectory prediction module fails to anticipate the movements of these mis-detected agents. 
To mitigate the issue of occlusions and detection limitations, one potential solution could be implementing cooperative perception by sharing detection information among agents using the vehicle-to-vehicle communication network \cite{yuan2022keypoints,xu2022opv2v}.

\section{Conclusion}
\label{conclusion}
In this paper, we propose a framework called ODTP that combines the perception module of the monocular Quasi-Dense 3D Object Tracking with the trajectory module of DCENet. 
This framework enables end-to-end detection, tracking, and prediction for autonomous driving.
We enhance the DCENet model by extending the dynamic maps to include agents' shape and pose information, which is termed DCENet++. 
This enhancement allows for more accurate mapping of interactions among agents.
Furthermore, we demonstrate that training the trajectory prediction module using multi-object tracking data helps the prediction module better adapt to cope with the noisy data perceived in real-world driving scenarios.
\newline
\newline
\noindent\textbf{Acknowledgements} This work is partially performed in the framework of project KaBa (Kamerabasierte Bewegungsanalyse aller Verkehrsteilnehmer für automatisiertes Fahren) supported by the European Regional Development Fund at VISCODA company.     

\bibliographystyle{IEEEtran}
\bibliography{bibliography}

\end{document}